\documentclass{article}

\usepackage[preprint]{neurips_2024}
\setcitestyle{numbers,square,comma}

\usepackage[utf8]{inputenc}
\usepackage[T1]{fontenc}
\usepackage{amsmath,amssymb,amsfonts}
\usepackage{bm}
\usepackage{graphicx}
\usepackage{booktabs}
\usepackage{multirow}
\usepackage{hyperref}
\hypersetup{
  colorlinks=true,
  linkcolor=blue!70!black,
  citecolor=blue!70!black,
  urlcolor=blue!70!black
}
\usepackage{cleveref}
\usepackage{float}
\usepackage{algorithm}
\usepackage{algpseudocode}
\usepackage{listings}
\usepackage{xcolor}
\usepackage{subcaption}
\usepackage{enumitem}
\usepackage{microtype}
\usepackage{tikz}
\usetikzlibrary{shapes,arrows,positioning,fit,backgrounds}

\title{Grounded Optimization: A Layered Engineering Framework for Reducing LLM Hallucination in Automated Personal Document Rewriting}

\author{
  Shashank Indukuri\thanks{Equal contribution.}\\
  \texttt{sinduku1@depaul.edu}
  \And
  Adarsh Agrawal\footnotemark[1]\\
  \texttt{adagrawal@cs.stonybrook.edu}
}

\begin{document}
\maketitle

\begin{abstract}
Large language models (LLMs) are increasingly applied to resume optimization for applicant
tracking systems, introducing hallucination failures distinct from general text generation:
anachronistic technology injection, cross-domain terminology contamination, structural
mutation, and content fabrication. We present \textbf{Grounded Optimization}, a five-layer
framework combining temporal context validation, deterministic contamination detection,
structural invariant enforcement, prompt-level grounding, and an evaluator agent.

In ablation experiments across three LLMs, four temperature settings, and six layer
configurations on 25 synthetic resumes spanning 14 industries, undefended baselines produce
2.48--5.36 detected hallucinations per resume. Among detectors independent of the active
defenses, temporal hallucinations are reduced by 50--95\% across all conditions; overall
detected hallucination rate falls to 0.04--0.24. Prompt-level grounding alone achieves zero
detected hallucinations at low temperature with a capable instruction-following model;
higher temperatures and weaker models reveal the need for the deterministic layers as a
complement. We release the contamination taxonomy, evaluation code, and raw data.
\end{abstract}

\section{Introduction}
\label{sec:intro}

The use of large language models for document optimization has grown rapidly, with resume tailoring representing one of the most commercially active applications. Services that rewrite resumes to improve alignment with job descriptions and ATS scoring algorithms now process large volumes of documents. Yet the hallucination behaviors documented in LLM general text generation~\citep{ji2023survey,zhang2023siren} manifest in particularly harmful ways when applied to personal documents:

\begin{enumerate}[leftmargin=*]
  \item \textbf{Temporal fabrication}: An LLM optimizing a 2018 software engineering role may inject references to LangChain (released late 2022) or Mixtral (released December 2023), creating verifiably false claims about the candidate's experience timeline.
  \item \textbf{Cross-domain contamination}: When rewriting a role at an AWS-centric company, the model may introduce Azure or GCP terminology to match job description keywords, adding multi-cloud terminology absent from the original role.
  \item \textbf{Structural mutation}: The model may silently merge, delete, or condense bullet points to reduce output length, removing genuine achievements in the process.
  \item \textbf{Content fabrication}: The model may invent company names, inflate metrics, or add certifications the candidate never earned.
\end{enumerate}

These failures carry concrete consequences: candidates may unknowingly submit resumes containing false claims, exposing them to disqualification or termination. Unlike hallucination in chatbots or search summaries, where users can verify outputs interactively, resume optimization typically operates in batch mode with minimal human review.

Hallucination mitigation has been studied extensively in open-domain question answering~\citep{manakul2023selfcheckgpt}, summarization~\citep{kryscinski2020evaluating}, and retrieval-augmented generation~\citep{lewis2020retrieval}. Prior work on hallucination in personal document optimization specifically is more limited. Concurrent system-level work has begun integrating anti-hallucination mechanisms into resume-tailoring pipelines (e.g., \citep{resumetailor2026}), but to our knowledge no published work characterizes the underlying hallucination modes as a taxonomy or systematically isolates the contribution of individual defense layers.

The ground truth in this domain is not an external knowledge base but the candidate's own career history, which the LLM receives as input and must \emph{enhance without distorting}.

We present \textbf{Grounded Optimization}, a five-layer defense-in-depth framework that addresses each hallucination mode through a distinct mechanism. The first two layers address the most common failures we observed: \textbf{temporal validation} (\Cref{sec:temporal}) prevents the model from injecting post-hoc technologies into historical roles by embedding release-date constraints in every prompt, and a \textbf{deterministic contamination detector} (\Cref{sec:contamination}) catches cloud-provider bleeding using a 257-service regex taxonomy without involving another LLM (which would introduce an additional hallucination surface). \textbf{Structural enforcement} (\Cref{sec:structural}) handles bullet compression: it counts roles and bullet points before and after optimization and rejects outputs that lose too much. \textbf{Prompt-level grounding} (\Cref{sec:prompt}) embeds explicit immutability rules for education, certifications, and company names directly in the agent prompts, providing a first-line defense before deterministic checks are applied. Finally, an \textbf{evaluator agent} (\Cref{sec:evaluator}) deploys a separate LLM instance as a quality gate that can reject and re-trigger the pipeline (partially independent; see \Cref{sec:eval-limitations} for an H2-specific coupling caveat).

Our framework is implemented as a multi-agent system built on LangGraph~\citep{langgraph2024} that processes resumes through five parallel specialized agents, each operating under the full defense stack. The system includes a fallback-merge mechanism that combines the best LLM output with preserved originals to retain all original content (\Cref{sec:system}).

The contributions of this paper are:
\begin{enumerate}[leftmargin=*]
  \item A \textbf{taxonomy of hallucination modes} specific to personal document optimization, distinguishing temporal, cross-domain, structural, and content fabrication failures (\Cref{sec:taxonomy}).
  \item A \textbf{five-layer engineering framework} combining deterministic validation, prompt engineering, and multi-agent adversarial checking, implemented and evaluated as a functional multi-agent system (\Cref{sec:framework}).
  \item A \textbf{deterministic cloud-provider contamination detector} covering 257 services across AWS, GCP, Azure, and on-premise stacks with two-tier confidence scoring (\Cref{sec:contamination}).
  \item An \textbf{ablation and sensitivity analysis} across 16 experimental conditions (three LLMs, four temperatures, six layer configurations, 680 LLM invocations) characterizing per-layer contributions, with documented evaluation limitations (\Cref{sec:evaluation}, \Cref{sec:eval-limitations}).
\end{enumerate}

\section{Hallucination Taxonomy for Personal Documents}
\label{sec:taxonomy}

We identify four distinct hallucination modes in personal document optimization, each with unique detection requirements and consequences.

\subsection{Temporal Fabrication (H1)}
The LLM inserts references to technologies that did not exist during the claimed time period. In the technology sector, where new tools emerge rapidly and carry strong ATS keyword signals, this is a frequent failure mode in our experiments. A role from January 2019 to March 2021 gets rewritten to include ``Implemented RAG pipelines using LangChain and vector databases,'' despite LangChain's release in late 2022 and the RAG paradigm~\citep{lewis2020retrieval}, introduced in 2020 but widely adopted starting in late 2022. We attribute this to a training-data artifact: the model has no mechanism to learn which tools existed in which year relative to a particular person's employment dates.

\subsection{Cross-Domain Contamination (H2)}
Cross-domain contamination proved to be the dominant failure mode in our experiments (79--89\% of baseline incidents). The model introduces terminology from a technology ecosystem not present in the original role: an AWS-focused position acquires Azure or GCP references because the job description mentions multi-cloud. In one test, a role at an AWS-only company (``Managed data pipelines using AWS Glue and Athena'') was rewritten as ``Orchestrated ETL workflows using Azure Data Factory and Synapse Analytics''---introducing Azure terminology absent from the original role. The model treats cloud services as interchangeable synonyms when optimizing for keyword coverage and has no awareness of organizational technology constraints.

\subsection{Structural Mutation (H3)}
Structural mutation is a subtler failure mode in which the model does not fabricate information but instead abbreviates it. A role with 8 bullet points may return with 4 or 5 ``enhanced'' entries that cover similar ground at a higher level of abstraction, while the most distinctive accomplishments---those that differentiate one candidate from another---are silently folded into generic summaries such as ``Maintained and optimized production systems.'' Unlike the other hallucination modes, structural mutation \emph{removes} truth rather than adding falsehood, making it harder to detect through surface-level review. The root cause appears to be that LLMs internalize conciseness as a quality signal, causing ``optimize'' to become ``condense'' without explicit instruction.

\subsection{Content Fabrication (H4)}
Content fabrication is the most straightforward failure mode: the model invents concrete details such as fabricated company names, inflated metrics (``Reduced API latency by 90\%'' in a role that mentioned no performance numbers), and non-existent certifications. This occurs less frequently than contamination or temporal fabrication in our data but is the hardest to detect post-hoc, as fabricated metrics resemble plausible candidate accomplishments and require access to the candidate's actual work history to verify.

\section{Defense Framework}
\label{sec:framework}

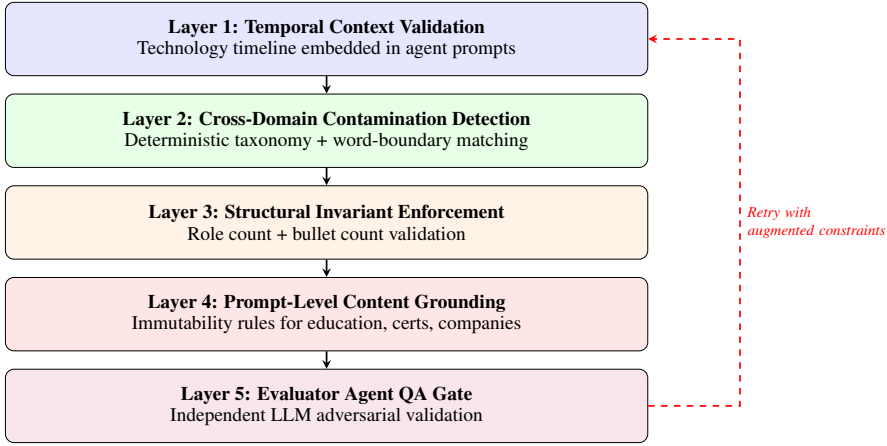
\begin{figure}[!htbp]
\centering
\resizebox{\textwidth}{!}{\begin{tikzpicture}[
  layer/.style={rectangle, draw, rounded corners, minimum width=10.5cm, minimum height=1.2cm, align=center, font=\small},
  arrow/.style={->, thick, >=stealth},
  note/.style={font=\scriptsize\itshape, text width=5cm}
]
  \node[layer, fill=blue!10] (l1) at (0, 5) {\textbf{Layer 1: Temporal Context Validation} \\ Technology timeline embedded in agent prompts};
  \node[layer, fill=green!10] (l2) at (0, 3.5) {\textbf{Layer 2: Cross-Domain Contamination Detection} \\ Deterministic taxonomy + word-boundary matching};
  \node[layer, fill=orange!10] (l3) at (0, 2) {\textbf{Layer 3: Structural Invariant Enforcement} \\ Role count + bullet count validation};
  \node[layer, fill=red!10] (l4) at (0, 0.5) {\textbf{Layer 4: Prompt-Level Content Grounding} \\ Immutability rules for education, certs, companies};
  \node[layer, fill=purple!10] (l5) at (0, -1) {\textbf{Layer 5: Evaluator Agent QA Gate} \\ Independent LLM adversarial validation};

  \draw[arrow] (l1) -- (l2);
  \draw[arrow] (l2) -- (l3);
  \draw[arrow] (l3) -- (l4);
  \draw[arrow] (l4) -- (l5);

  \draw[arrow, dashed, color=red] (l5.east) -- ++(1.5, 0) |- node[right, note, pos=0.25] {Retry with\\augmented constraints} (l1.east);
\end{tikzpicture}}
\caption{Five-layer defense-in-depth architecture. Each layer addresses a distinct hallucination mode. Failed validation at Layer 5 triggers a retry cycle with contamination warnings and structural constraints injected into the prompt. After 3 failed retries, the system falls back to a merge of the best LLM output with original content.}
\label{fig:architecture}
\end{figure}

Our layered defense addresses each hallucination mode through a distinct layer, as shown in \Cref{fig:architecture}. Two of the five layers operate at generation time: Layer~4 embeds immutability constraints directly in the agent prompts \emph{before} the LLM call, making it the first defense to act on any given optimization cycle. Layers~1--3 and~5 operate post-generation, validating and potentially reverting the LLM's output before it is accepted. The layers are numbered by their role in the validation pipeline; the execution order within a single cycle is L4 (prompt injection) $\to$ LLM call $\to$ L1--L3 (output validation) $\to$ L5 (evaluator gate). Failures at any post-generation layer trigger retry with augmented constraints or fallback to original content.

\subsection{Layer 1: Temporal Context Validation}
\label{sec:temporal}

The temporal context layer prevents anachronistic technology injection (H1) by building a per-resume timeline and embedding it as a constraint in every agent prompt.

Given a resume $R$ with experience entries $E = \{e_1, \ldots, e_n\}$, each with start/end dates, we construct a temporal context $\text{TC}(R)$ containing the career span, a technology-to-year-range mapping derived from bullet-point scanning, and the current year. We maintain a curated mapping of technology release dates (e.g., LangChain$\to$2022, Vertex AI$\to$2021) that constrain which technologies may appear in which roles. The full timeline construction algorithm and release-date table are in \Cref{app:temporal}.

The temporal context is serialized and injected into every agent prompt, instructing the LLM to verify technology existence during each role's time period.

\subsection{Layer 2: Cross-Domain Contamination Detection}
\label{sec:contamination}

The contamination detection layer addresses cross-domain bleeding (H2) through a fully deterministic, LLM-free mechanism. An initial LLM-based approach---asking the model to verify its own output for foreign cloud services---proved functional but added latency and cost per invocation. Because cloud service names form a finite, enumerable set, a deterministic regex-based approach is both sufficient and more efficient.

We construct a taxonomy $\mathcal{T}$ of 257 cloud services across four ecosystems (AWS: 76, GCP: 53, Azure: 64, On-Premise: 64), plus a cloud-agnostic set of 69 provider-independent technologies. Each ecosystem entry consists of explicit provider keywords (e.g., ``aws'') and service names (e.g., ``sagemaker''). Detection uses two-tier word-boundary regex matching: Tier~1 attributes on a single explicit-keyword match; Tier~2 requires $\geq$2 service-name matches to handle ambiguity (e.g., ``lambda'' as AWS Lambda vs.\ the Python keyword). The full detection algorithm and ambiguity resolution are in \Cref{app:detection}.

The key design decision: we compare each role's cloud signature \emph{before and after} optimization, flagging only \emph{newly introduced} providers:
\begin{equation}
\text{Contaminated}(e_i) = \text{Clouds}(e_i^{\text{updated}}) \setminus \text{Clouds}(e_i^{\text{original}}) \neq \emptyset
\end{equation}
When contamination is detected, the role's responsibilities are reverted to originals, a contamination warning is injected into the retry prompt, and optimization is retried with augmented constraints.

\subsection{Layer 3: Structural Invariant Enforcement}
\label{sec:structural}

Structural mutation (H3) is addressed through pre/post counting of semantic units with tolerance-aware validation.

Before optimization, we record a structural signature $\text{Sig}(R) = (|E|, \{|b_i|\})$, the number of experience entries and bullet counts per entry. After optimization, we validate that $|E'| \geq |E|$ and $|b'_i| \geq |b_i| - 1$ for each entry, accommodating minor restructuring while preventing significant content loss. When validation fails, the retry prompt includes explicit structural targets. After 3 failed attempts, a deterministic fallback merge ensures all original content is retained (\Cref{app:structural}).

\subsection{Layer 4: Prompt-Level Content Grounding}
\label{sec:prompt}

Content fabrication (H4) is addressed through explicit immutability declarations embedded in every agent prompt. While prompt-level constraints alone are insufficient at higher temperatures and on weaker models (Experiments~2--3), they serve as a strong first line of defense that significantly reduces the frequency of violations the subsequent layers must catch.

The grounding constraints are organized into four categories:

\begin{enumerate}[leftmargin=*]
  \item \textbf{Content Preservation}: ``Preserve the exact number of bullet points for each entry. DO NOT reduce or condense them.''
  \item \textbf{Factual Immutability}: ``DO NOT hallucinate, add, or modify educational details (institution name, location, degree information).''
  \item \textbf{Entity Integrity}: ``DO NOT create a new company or use placeholder names.''
  \item \textbf{Metric Realism}: ``Ensure metrics and numbers are realistic for the time period.''
\end{enumerate}

\subsection{Layer 5: Evaluator Agent QA Gate}
\label{sec:evaluator}

The final layer deploys an independent LLM instance as an adversarial quality-control agent, implementing a generator-critic architecture~\citep{gou2024critic} specialized for personal document validation.

The evaluator receives the original resume, the rewritten resume, and the target job description, and returns $(\text{is\_acceptable} \in \{0,1\}, \text{feedback})$. It checks for content removal, JD alignment, and plausible ATS improvement. As a distinct model instance, it avoids generator-bias transfer; on rejection, feedback is injected into the next rewrite attempt. If the evaluator itself fails (timeout or malformed output), it defaults to rejection rather than silently accepting the candidate output.

\subsection{Implementation}
\label{sec:system}

The framework is implemented as a multi-agent pipeline on LangGraph~\citep{langgraph2024}. The system processes resumes through four stages (parse, score, rewrite, re-score) with up to 5 optimization cycles. Five specialized agents (Summary, Skills, Experience, Projects, Education) run in parallel; the Experience Agent receives the full defense stack because professional experience is where most hallucinations occur. A LangGraph \texttt{AgentState} preserves original data alongside optimized versions throughout, enabling fallback merge at any point. Full pipeline details are in \Cref{app:system}.

\section{Evaluation}
\label{sec:evaluation}

We evaluate the framework through three complementary experiments following evaluation methodology from recent hallucination benchmarks~\citep{min2023factscore,li2023halueval}: (1)~an ablation study measuring each defense layer's contribution, (2)~a multi-model generalization study across three LLMs, and (3)~a temperature sensitivity analysis. All experiments use 25 synthetic resumes, 42 roles, 188 bullet points, and 5 job descriptions, with seed=42 for reproducibility.

\subsection{Dataset}
We construct a corpus of 25 synthetic resumes spanning 14 industries (technology, finance, healthcare, manufacturing, consulting, retail, education, energy, government, media, logistics, telecom, insurance, and real estate). Resumes contain 42 professional roles totaling 188 bullet points, ranging from 1 to 6 roles per resume and covering career histories from 2013 to 2026. Five adversarial job descriptions are designed to induce hallucination: a multi-cloud AI position requiring both AWS and Azure, a GCP ML role requesting RAG experience, an AWS full-stack role mentioning generative AI, an Azure data analytics role, and a generic senior role. Each resume is paired with one job description in round-robin assignment.

\subsection{Evaluation Protocol}
For each experiment, every resume--JD pair is processed under the specified configuration and the output is evaluated by four deterministic hallucination detectors (H1--H4) that compare each optimized role against its original:
\begin{enumerate}[leftmargin=*]
  \item \textbf{H1 Temporal detector}: Checks for technologies released after the role's end date, using a curated mapping of technology release years.
  \item \textbf{H2 Contamination detector}: Uses the cloud-provider taxonomy (\Cref{sec:contamination}) to identify newly introduced cloud services not in the original role.
  \item \textbf{H3 Structural detector}: Compares bullet-point counts, flagging any loss of $>$1 bullet.
  \item \textbf{H4 Fabrication detector}: Checks for company name changes and title mutations exceeding 50\% word overlap.
\end{enumerate}

\noindent\textbf{Known detector--defense coupling (H2).} The H2 detector and the Layer~2 defense share the same underlying \texttt{detect\_role\_contamination} function from the cloud taxonomy module. When Layer~2 is active in a configuration, any contamination it detects is reverted \emph{before} the H2 detector evaluates the output; the detector and the defense therefore cannot disagree by construction. H2 counts in L2-active configurations are thus a tautological consequence of L2's revert behavior and should not be interpreted as independent empirical measurements. We retain these counts in the tables for completeness but discuss the implication in \Cref{sec:eval-limitations} and treat them accordingly when interpreting results.

\subsection{Metrics}
We report \textbf{Hallucination Rate (HR)}: mean detected hallucination incidents per resume, with standard deviation ($\sigma$) and 95\% confidence interval.

\subsection{Experiment 1: Ablation Study}

We test six defense configurations with GPT-4.1-nano at temperature=0 to isolate each layer's contribution:

\begin{table}[!htbp]
\centering
\resizebox{\textwidth}{!}{\begin{tabular}{lccccccl}
\toprule
 & & & & \multicolumn{4}{c}{\textbf{Detected Incidents by Type}} \\
\cmidrule(lr){5-8}
\textbf{Defense Configuration} & \textbf{Detect. Rate$\downarrow$} & \textbf{Std Dev} & \textbf{95\% CI} & \textbf{Temporal} & \textbf{Contam.$^\dagger$} & \textbf{Structural} & \textbf{Fabrication} \\
\midrule
No defense (baseline) & 2.48 & 3.84 & $\pm$1.50 & 7 & 55 & 0 & 0 \\
L4 only (prompt grounding) & 0.00 & 0.00 & $\pm$0.00 & 0 & 0 & 0 & 0 \\
L1+L4 (+ temporal) & 0.12 & 0.43 & $\pm$0.17 & 2 & 0 & 0 & 1 \\
L1+L2+L4 (+ contamination) & 0.08 & 0.27 & $\pm$0.11 & 1 & 0 & 1 & 0 \\
L1+L2+L3+L4 (+ structural) & 0.16 & 0.37 & $\pm$0.14 & 1 & 0 & 0 & 3 \\
\textbf{Full (L1+L2+L3+L4+L5)} & \textbf{0.12} & \textbf{0.33} & $\bm{\pm}$\textbf{0.13} & \textbf{1} & \textbf{0} & \textbf{0} & \textbf{2} \\
\bottomrule
\multicolumn{8}{l}{\footnotesize $^\dagger$Contamination counts under L2-active configs are mechanically zero by construction (\Cref{sec:eval-limitations}).} \\
\end{tabular}}
\caption{Ablation study on 25 resumes (GPT-4.1-nano, $t$=0). Contamination ($\dagger$) counts when Layer~2 is active are mechanically zero by construction (see \Cref{sec:eval-limitations}) and are not independent measurements. The undefended baseline produces 62 detected hallucination incidents (2.48 per resume). Prompt-level grounding alone (L4) achieves zero detected hallucinations in this single (model, temperature) configuration; the temperature and multi-model experiments demonstrate this does not generalize.}
\label{tab:ablation}
\end{table}

\textbf{Observation:} The L4-only result is informative: with a strong instruction-following model at $t$=0, prompt grounding alone produces zero detected hallucinations. This is consistent with the view that modern LLMs can respect explicit behavioral constraints at low temperature. Experiments~2 and~3 show the result does not generalize across models or temperatures. L4-only (HR=0.00) also outperforms the Full framework (HR=0.12) at this single configuration. Inspection of the 3 residual incidents under Full reveals 1~H1 (a 2019 role received a post-2022 technology reference despite Layer~1 constraints) and 2~H4 (title reformulations such as ``Senior Data Analyst''$\to$``Lead Data Analyst'' crossing the 50\% word-overlap threshold). These H4 cases are likely false positives of our coarse fabrication detector. Excluding them, Full achieves HR=0.04, consistent with L4-only. The L4-vs-Full gap is therefore most likely an artifact of H4 detector sensitivity rather than evidence that additional layers harm performance.

\subsection{Experiment 2: Multi-Model Generalization}

We test the baseline and full framework across three LLMs of varying capability at $t$=0:

\begin{table}[!htbp]
\centering
\resizebox{\textwidth}{!}{\begin{tabular}{llccccccc}
\toprule
 & & & & \multicolumn{4}{c}{\textbf{Detected Incidents by Type}} & \\
\cmidrule(lr){5-8}
\textbf{Model} & \textbf{Defense} & \textbf{Detect. Rate$\downarrow$} & \textbf{Std Dev} & \textbf{Temporal} & \textbf{Contam.$^\dagger$} & \textbf{Structural} & \textbf{Fabrication} & \textbf{Reduction (\%)$^\dagger$} \\
\midrule
GPT-4.1-nano & Baseline & 2.48 & 3.84 & 7 & 55 & 0 & 0 & --- \\
GPT-4.1-nano & Full & 0.12 & 0.33 & 1 & 0 & 0 & 2 & 95.2 \\
\midrule
GPT-4o-mini & Baseline & 5.36 & 5.62 & 20 & 106 & 0 & 8 & --- \\
GPT-4o-mini & Full & 0.04 & 0.20 & 1 & 0 & 0 & 0 & \textbf{99.3} \\
\midrule
Llama-3.1-8B & Baseline & 4.44 & 5.61 & 19 & 88 & 0 & 4 & --- \\
Llama-3.1-8B & Full & 0.12 & 0.33 & 1 & 0 & 0 & 2 & 97.3 \\
\bottomrule
\multicolumn{9}{l}{\footnotesize $^\dagger$Contamination and reduction figures inherit the H2 detector--defense coupling (\Cref{sec:eval-limitations}).} \\
\end{tabular}}
\caption{Multi-model evaluation at $t$=0. Contamination ($\dagger$) counts under the Full configuration are mechanically zero by construction (see \Cref{sec:eval-limitations}). Less capable models produce 2--4$\times$ more baseline detected hallucinations. Reduction percentages are computed against detected-HR and inherit the H2 caveat; we report them for engineering reference but do not claim elimination in an independent-evaluator sense.}
\label{tab:multimodel}
\end{table}

\textbf{Observation:} The framework is applicable across model families (OpenAI, Meta/Groq). GPT-4o-mini produces 2.2$\times$ more baseline detected hallucinations than GPT-4.1-nano, and Llama-3.1-8B 1.8$\times$ more. Under the Full configuration, detected-HR falls to near zero on our current metrics. H2 counts of zero under Full are structurally guaranteed (\Cref{sec:eval-limitations}); the non-tautological observations are (i) the large baseline H2 counts across all three models, which show the defense target is real, and (ii) the reductions in H1, H3, and H4 which are measured by detectors distinct from any active defense component.

\subsection{Experiment 3: Temperature Sensitivity}

We vary the sampling temperature from 0 to 1.0 with GPT-4.1-nano:

\begin{table}[!htbp]
\centering
\resizebox{\textwidth}{!}{\begin{tabular}{ccccccccr}
\toprule
 & & & & \multicolumn{4}{c}{\textbf{Detected Incidents by Type}} & \\
\cmidrule(lr){5-8}
\textbf{Temp.} & \textbf{Defense} & \textbf{Detect. Rate$\downarrow$} & \textbf{Std Dev} & \textbf{Temporal} & \textbf{Contam.$^\dagger$} & \textbf{Structural} & \textbf{Fabrication} & \textbf{Reduction (\%)$^\dagger$} \\
\midrule
0.0 & Baseline & 2.48 & 3.84 & 7 & 55 & 0 & 0 & --- \\
0.0 & Full & 0.12 & 0.33 & 1 & 0 & 0 & 2 & 95.2 \\
\midrule
0.3 & Baseline & 2.12 & 2.89 & 7 & 45 & 0 & 1 & --- \\
0.3 & Full & 0.16 & 0.46 & 1 & 0 & 0 & 3 & 92.5 \\
\midrule
0.7 & Baseline & 1.72 & 2.91 & 8 & 34 & 0 & 1 & --- \\
0.7 & Full & 0.16 & 0.37 & 2 & 0 & 0 & 2 & 90.7 \\
\midrule
1.0 & Baseline & 1.80 & 3.63 & 8 & 36 & 0 & 1 & --- \\
1.0 & Full & 0.24 & 0.43 & 4 & 0 & 0 & 2 & 86.7 \\
\bottomrule
\multicolumn{9}{l}{\footnotesize $^\dagger$Contamination and reduction figures inherit the H2 detector--defense coupling (\Cref{sec:eval-limitations}).} \\
\end{tabular}}
\caption{Temperature sensitivity (GPT-4.1-nano). Contamination ($\dagger$) counts under Full are mechanically zero by construction (see \Cref{sec:eval-limitations}). Baseline detected-hallucinations decrease slightly at higher temperatures in this data; we are cautious interpreting this trend given $\sigma > \mu$ on all baselines. Residual violations under Full are H1 (temporal) and H4 (minor fabrication); H1 residuals grow from 1 at $t$=0 to 4 at $t$=1.0, consistent with reduced prompt compliance under higher stochasticity.}
\label{tab:temperature}
\end{table}

\textbf{Observation:} Detected-HR under Full increases from 0.12 at $t$=0 to 0.24 at $t$=1.0, driven almost entirely by H1 (temporal) residuals (1 $\to$ 4) which are measured by a detector independent of any active defense layer. The H1 trend is the most interpretable signal in this experiment because it is free of the detector--defense coupling that affects H2. The graceful degradation of H1 detected-count under increasing stochasticity suggests prompt compliance weakens with temperature, motivating deterministic layers as a complement rather than a replacement for prompt-based grounding.

\subsection{Cross-Experiment Summary}

Taken together, the three experiments reveal a clear interaction: prompt-level grounding (L4) is sufficient at low temperature with a strong model, but its effectiveness degrades predictably with both increasing temperature (H1 residuals: $1\to4$) and decreasing model capability (baseline HR: $2.48\to5.36$ across models). The deterministic layers (L1--L3) provide the most value precisely where prompt compliance is weakest --- high temperature and weaker models --- rather than as a uniform improvement over prompt grounding alone. The remaining residuals under defended configurations are almost entirely H1 (temporal) and likely-false-positive H4 (minor title reformulations), suggesting that further gains require either a stronger temporal enforcement mechanism or a semantics-aware fabrication detector.

\section{Related Work}
\label{sec:related}

\paragraph{LLM hallucination.} Hallucination has been extensively studied across summarization, translation, and dialogue~\citep{ji2023survey,zhang2023siren,huang2023hallucination}. Detection methods include sampling consistency (SelfCheckGPT~\citep{manakul2023selfcheckgpt}) and fine-grained factuality scoring (FActScore~\citep{min2023factscore}). These approaches target \emph{general knowledge} hallucination where ground truth exists in external corpora. Personal document optimization is different in kind: the ground truth is the input document itself, and hallucination manifests as distortion of the user's own data.

\paragraph{Constrained generation and multi-agent systems.} Constrained decoding~\citep{hu2019improved,lu2021neurologic} enforces token-level constraints; our structural enforcement operates at the semantic-unit level (roles, bullet points). Multi-agent debate~\citep{du2023improving}, self-reflection~\citep{shinn2023reflexion}, and critic-generator frameworks~\citep{gou2024critic} improve LLM reliability through adversarial checking. Our evaluator agent extends this paradigm to personal document QA, where the critic checks content preservation rather than general quality. A related pattern appears in code generation, where deterministic catalog selection and access-control gating before LLM SQL generation reduce execution errors~\citep{indukuri2026schema}; both settings suggest deterministic layers around generation can complement prompt-level approaches.

\paragraph{Taxonomy-driven LLM evaluation.} Recent benchmarks structure LLM behavioral evaluation around explicit hazard or failure-mode taxonomies. The MLCommons AI Safety Benchmark v0.5~\citep{vidgen2024aisafety} introduces a 13-hazard taxonomy for general-purpose chat assistants. Our four-mode hallucination taxonomy (H1--H4) is narrower in scope but follows a similar methodological pattern: structured failure-mode definitions paired with category-specific detectors and per-category reporting.

\paragraph{Resume processing.} Prior work focuses on parsing and screening~\citep{sinha2021resume}, matching~\citep{deng2018job}, scoring~\citep{mittal2020resume}, and end-to-end LLM-based resume generation~\citep{resumeflow2024}. Recent concurrent work on resume-tailoring systems has begun incorporating anti-hallucination guardrails as a system component~\citep{resumetailor2026}. Our work differs in framing: rather than building a single tailored system, we characterize the hallucination behaviors specific to this domain as a taxonomy and isolate the empirical contribution of individual defense layers.

\section{Limitations and Discussion}
\label{sec:limitations}

\subsection{H2 Detector--Defense Coupling}
\label{sec:eval-limitations}
The H2 detector and Layer~2 defense share the same \texttt{detect\_role\_contamination} function. Because Layer~2 reverts contaminated output \emph{before} the detector runs, H2 counts under L2-active configurations are mechanically zero by construction; we flag these with $\dagger$ throughout. The large baseline H2 counts (measured without any active defense) confirm the defense target is real, but we cannot independently verify Layer~2 \emph{eliminates} contamination rather than merely hiding it from our own detector. An independent NLI-based evaluator on the existing 680 outputs is the highest-priority extension.

Other limitations are more conventional: we do not compare against SelfCheckGPT, FActScore, or CRITIC (our H1--H4 detectors are task-specific); per-resume counts are zero-inflated and heavy-tailed ($\sigma > \mu$ on all baselines); the 25-resume synthetic dataset and 257-service taxonomy miss real-world distributions and long-tail platforms (Salesforce, SAP); and the framework detects presence but not magnitude hallucinations.

\paragraph{Future work.} An independent NLI-based H2 evaluator on the existing 680 outputs is the highest-priority extension. Beyond that: comparison with external hallucination detectors (SelfCheckGPT, FActScore) and commodity guardrail frameworks (e.g., NeMo Guardrails, Guardrails AI) as alternative enforcement backends, human annotation on real resumes, and extension to other personal documents. All code, taxonomy, and raw data are available at \url{https://github.com/shashank-indukuri/grounded-optimization}.


\bibliographystyle{unsrtnat}
\bibliography{references}

\appendix

\section{Temporal Context Validation Details}
\label{app:temporal}

\subsection{Timeline Construction}

Given a resume $R$ with professional experience entries $E = \{e_1, \ldots, e_n\}$, where each entry $e_i$ has start date $s_i$ and end date $t_i$, we construct:

\begin{equation}
\text{TC}(R) = \left\{
  \begin{aligned}
    &\text{career\_span}: [s_{\min}, t_{\max}] \\
    &\text{tech\_timeline}: \{(\tau, [\text{first\_used}, \text{last\_used}])\} \\
    &\text{current\_year}: y_{\text{now}}
  \end{aligned}
\right\}
\end{equation}

where $\tau$ ranges over technologies mentioned in existing bullet points, and first/last used years are derived from the dates of roles containing $\tau$.

\subsection{Release Date Constraints}

\begin{table}[h]
\centering
\small
\begin{tabular}{lll}
\toprule
\textbf{Technology} & \textbf{Release Year} & \textbf{Constraint} \\
\midrule
LangChain & 2022 & Cannot appear in pre-2022 roles \\
LlamaIndex & 2022 & Cannot appear in pre-2022 roles \\
Vertex AI & 2021 & Cannot appear in pre-2021 roles \\
Mixtral & 2023 & Cannot appear in pre-2023 roles \\
RAG (paradigm) & 2022 & Cannot appear in pre-2022 roles \\
\bottomrule
\end{tabular}
\caption{Example technology release date constraints embedded in temporal context.}
\label{tab:temporal}
\end{table}

\subsection{Timeline Construction Algorithm}

\begin{algorithm}[H]
\caption{Temporal Context Construction}
\label{alg:temporal}
\begin{algorithmic}[1]
\Require Resume $R$ with experience entries $E$
\Ensure Temporal context $\text{TC}(R)$
\State $\text{total\_months} \gets 0$
\State $\text{tech\_timeline} \gets \{\}$
\For{$e_i \in E$}
  \State Parse $s_i, t_i$ from $e_i.\text{dates}$
  \State $\text{total\_months} \mathrel{+}= (t_i - s_i)$ in months
  \For{responsibility $r \in e_i.\text{bullets}$}
    \For{technology $\tau$ detected in $r$}
      \State Update $\text{tech\_timeline}[\tau].\text{first\_used}$
      \State Update $\text{tech\_timeline}[\tau].\text{last\_used}$
    \EndFor
  \EndFor
\EndFor
\State \Return $\text{TC}(R) = \{\text{total\_months}, \text{tech\_timeline}, y_{\text{now}}\}$
\end{algorithmic}
\end{algorithm}

\section{Recency-Bounded Optimization}
\label{app:recency}

An additional grounding mechanism limits the optimization scope: only roles within the most recent 7 years are processed. The 7-year threshold was determined empirically: beyond this point, the marginal benefit of optimization diminished while hallucination risk for older roles increased. Older roles are passed through untouched, removing them from the optimization scope and therefore from this framework's hallucination risk.

\begin{equation}
E_{\text{process}} = \{e_i \in E : \text{months}(e_i) \leq 84\}, \quad E_{\text{preserve}} = E \setminus E_{\text{process}}
\end{equation}

The recency split operates chronologically from the most recent role, accumulating months until the 7-year threshold is reached.

\section{Contamination Detection Details}
\label{app:detection}

\subsection{Two-Tier Detection Algorithm}

\begin{algorithm}[H]
\caption{Two-Tier Cloud Provider Detection}
\label{alg:detect}
\begin{algorithmic}[1]
\Require Text $x$, Taxonomy $\mathcal{T}$
\Ensure Detected providers $P$
\State $P \gets \emptyset$
\For{$(j, K_j, S_j) \in \mathcal{T}$}
  \If{$\exists k \in K_j : \textsc{WordMatch}(k, x)$}
    \State $P \gets P \cup \{j\}$; \textbf{continue}
  \EndIf
  \State $m \gets |\{s \in S_j : \textsc{WordMatch}(s, x)\}|$
  \If{$m \geq 2$}
    \State $P \gets P \cup \{j\}$
  \EndIf
\EndFor
\State \Return $P$ if $P \neq \emptyset$ else $\{\text{Cloud-Agnostic}\}$
\end{algorithmic}
\end{algorithm}

The $\textsc{WordMatch}$ function uses compiled word-boundary regex patterns ($\backslash$b\textit{term}$\backslash$b) with case-insensitive matching, ensuring that ``S3'' matches the AWS service but not substrings like ``MS365.''

\subsection{Ambiguity Resolution}

Certain terms require contextual disambiguation. For example, ``Glue'' could refer to AWS Glue (an ETL service) or general adhesive. We handle ambiguous terms by requiring co-occurrence with provider context:

\begin{lstlisting}[caption={Ambiguity resolution for context-dependent terms},label=lst:ambiguity]
if service in ["glue", "power bi", "databricks"]:
    if service == "glue" and "aws" not in text_lower:
        continue  # Not AWS Glue without AWS context
    if service == "databricks" and provider == "Azure" \
       and "azure" not in text_lower:
        continue  # Not Azure Databricks without context
\end{lstlisting}

\section{Structural Enforcement Details}
\label{app:structural}

\subsection{Retry Prompt Template}

When structural validation fails, the retry prompt includes explicit targets:

\begin{lstlisting}[caption={Structural retry injection},label=lst:retry]
CRITICAL RETRY INSTRUCTION - ATTEMPT {attempt}:
You MUST return EXACTLY {role_count} roles:
- role_0: Senior Engineer at CompanyA: 6 bullets
- role_1: Engineer at CompanyB: 5 bullets
DO NOT skip any roles. Process ALL roles from input.
\end{lstlisting}

\subsection{Fallback Merge Algorithm}

After 3 failed validation attempts, the system executes a deterministic merge:

\begin{algorithm}[H]
\caption{Fallback Merge Strategy}
\label{alg:fallback}
\begin{algorithmic}[1]
\Require Original roles $E$, Best LLM output roles $E'$
\Ensure Merged roles $M$ with zero content loss
\State $\text{map} \gets \{(\text{title}_i, \text{company}_i) \to e'_i : e'_i \in E'\}$
\State $M \gets []$
\For{$e_i \in E$}
  \State $\text{key} \gets (\text{title}_i, \text{company}_i)$
  \If{$\text{key} \in \text{map}$}
    \State $e'_i \gets \text{map}[\text{key}]$
    \If{$|b'_i| < |b_i|$}
      \State $b'_i \gets b'_i + b_i[|b'_i|:]$
    \EndIf
    \State $M.\text{append}(e'_i)$
  \Else
    \State $M.\text{append}(e_i)$
  \EndIf
\EndFor
\State \Return $M$
\end{algorithmic}
\end{algorithm}

By construction, no role or bullet point can be dropped by this merge, even when the LLM consistently fails structural validation.

\section{System Architecture Details}
\label{app:system}

\subsection{Pipeline Stages}

The system processes resumes through a four-stage state machine:

\begin{enumerate}[leftmargin=*]
  \item \textbf{Parse}: LLM-based PDF-to-JSON conversion, producing structured resume data with typed fields (contact info, experience entries with dates and bullet points, education, skills, projects, certifications).
  \item \textbf{Score}: ATS scoring against the target job description, producing section-level feedback and an aggregate score.
  \item \textbf{Rewrite}: Multi-agent parallel optimization with all five defense layers active.
  \item \textbf{Re-Score}: The optimized resume is scored again; if the score has not improved sufficiently, the rewrite stage is repeated (up to 5 cycles).
\end{enumerate}

\subsection{Agent Specialization}

Five specialized agents run in parallel:

\begin{itemize}[leftmargin=*]
  \item \textbf{Summary Agent}: Optimizes the professional summary
  \item \textbf{Skills Agent}: Aligns skills with job requirements
  \item \textbf{Experience Agent}: Rewrites professional experience (full defense stack)
  \item \textbf{Projects Agent}: Enhances project descriptions
  \item \textbf{Education Agent}: Validates (but does not modify) education entries
\end{itemize}

The Experience Agent receives the heaviest defense treatment because professional experience is where most hallucinations occur. It alone executes the retry-validation-fallback loop described in \Cref{sec:structural}.

\subsection{State Management}

The system maintains a LangGraph \texttt{AgentState} that carries the resume through all stages, preserving the original data alongside optimized versions. This enables the fallback merge at any point and provides full diff-based auditability of every change.

\section{Qualitative Examples}
\label{app:qualitative}

\textbf{Cross-model contamination (GPT-4o-mini)}: Optimizing a GCP-only ML Engineer role for a multi-cloud JD, GPT-4o-mini injected 7 Azure terms (``Azure ML Studio,'' ``Cosmos DB,'' ``Azure Monitor'') and 4 AWS terms (``SageMaker,'' ``Glue,'' ``Redshift,'' ``CloudWatch'') into a single role, introducing AWS and Azure terminology absent from the original GCP-only role. The deterministic detector identified all 11 foreign terms and reverted the output.

\textbf{Temporal fabrication at high temperature}: At $t$=1.0, GPT-4.1-nano rewrote a 2017--2019 bank analyst role to include ``leveraged vector databases for semantic search,'' a technology paradigm that emerged in 2022. At $t$=0, the same model respected the temporal constraint. This demonstrates the temperature-dependent reliability of prompt-level constraints.

\textbf{Intern hallucination across models}: A software intern (Python Flask, pytest, 3 months) was optimized for a GCP ML Engineer position. All three models injected cloud services (GCP: ``Cloud Run,'' ``Vertex AI''; AWS: ``Lambda,'' ``SageMaker'') at baseline, fabricating cloud expertise for a candidate with zero cloud experience. The framework correctly identified and reverted all contamination.

\end{document}